\pdfoutput=1

\documentclass[11pt]{article}

\usepackage[final]{acl}

\usepackage{authblk}
\usepackage{lipsum}

\newcommand\blfootnote[1]{%
  \begingroup
  \renewcommand\thefootnote{}\footnote{#1}%
  \addtocounter{footnote}{-1}%
  \endgroup
}

\usepackage{times}
\usepackage{latexsym}

\usepackage[T1]{fontenc}

\usepackage[utf8]{inputenc}

\usepackage{microtype}

\usepackage{inconsolata}

\usepackage{graphicx}

\usepackage{multirow}
\usepackage{expex}

\title{Dialogues of Dissent: Thematic and Rhetorical Dimensions of Hate and Counter-Hate Speech in Social Media Conversations}

\author[1$\dagger$]{Effi Levi}
\author[2$\dagger$]{Gal Ron}
\author[2]{Odelia Oshri}
\author[2]{Shaul R. Shenhav}
\affil[1]{Department of Computer Science, The Hebrew University of Jerusalem}
\affil[2]{Department of Political Science, The Hebrew University of Jerusalem}
\affil[ ]{{\tt efle@cs.huji.ac.il}}
\affil[ ]{{\tt \{gal.ron2$|$odelia.oshri$|$shaul.shenhav\}@mail.huji.ac.il}}

\begin{document}
\maketitle

\blfootnote{$^\dagger$Both authors contributed equally to this work.}

\begin{abstract}

We introduce a novel multi-labeled scheme for joint annotation of hate and counter-hate speech in social media conversations, categorizing hate and counter-hate messages into thematic and rhetorical dimensions. The thematic categories outline different discursive aspects of each type of speech, while the rhetorical dimension captures how hate and counter messages are communicated, drawing on Aristotle's Logos, Ethos and Pathos. We annotate a sample of 92 conversations, consisting of 720 tweets, and conduct statistical analyses, incorporating public metrics, to explore patterns of interaction between the thematic and rhetorical dimensions within and between hate and counter-hate speech. Our findings provide insights into the spread of hate messages on social media, the strategies used to counter them, and their potential impact on online behavior.

\end{abstract}

\section{Introduction}

Social media platforms enable instant communication and allow for a vast sharing of ideas, but they have also become a main arena for the spread of hateful discourse (see~\citet{elsherief2018peer,cervone2021language}).
Much research has been devoted to studying the factors that contribute to the spread of online hate~\cite{ben2016hate,mathew2019spread}, as well as to automatically detecting hate speech~\cite{waseem-hovy-2016-hateful,mathew2021hatexplain} and to classifying it into different types of hateful and abusive language~\cite{nobata2016abusive,waseem-etal-2017-understanding,founta2018large,elsherief-etal-2021-latent,vidgen-etal-2021-learning,ron-etal-2023-factoring}.

Side-by-side with online hate, social media has seen the emergence of \textit{counter-hate speech}, which may be defined as \textit{speech that constitutes a direct reference or response to hate speech that opposes or denounces the message and/or the addresser, or the addresser's social group}~\cite{Benesch2014CounteringDS,wright-etal-2017-vectors}.
Counter-hate messages seek to interrupt the spread of hate by challenging it (whether explicitly or implicitly), thus attempting to change the discourse from within. Research in this field has engaged in identifying different counter hate speech strategies~\cite{benesch2016counterspeech,mathew2019thou},
and exploring their efficacy and impact on those involved in, or exposed to the hateful communication~\cite{benesch2016considerations,hangartner2021empathy}.
Much effort has additionally focused on the generation of counter-hate speech arguments
~\cite{qian-etal-2019-benchmark,chung-etal-2019-conan,sahacountergedi}, 
and some works have also examined the relationships and interactions between the two types of speech within organic online conversations
~\cite{mathew2020interaction,albanyan2022pinpointing,albanyan-etal-2023-counterhate}. 

Among the studies that engage in hate or counter-hate speech annotation, several have also addressed the forms of speech through which the message is communicated. For example, \citet{furman-etal-2023-argumentative} encoded argumentative components of online hate speech, and \citet{baider2023accountability} incorporated argumentative and affective strategies into an annotation scheme for counter-hate speech.

In this work, we introduce a novel multi-labeled scheme for joint annotation of hate and counter-hate speech in social media conversations, while taking the conversational context into account. The annotation scheme encodes hate and counter-hate along two distinct dimensions: a \textit{thematic} dimension, which factors hate and counter-hate into different discursive aspects, and a \textit{rhetorical} dimension, which captures the way in which hate and counter-hate are communicated to the reader.

Our work offers a holistic approach for encoding hate as well as counter-hate speech, on both the thematic dimension and the rhetorical one.
The thematic dimension is defined separately for each of the encoded phenomena (hate/counter-hate). For hate, it is designed to capture the ways in which the target of hate is perceived and addressed,
through the following three categories: \textit{contemptuous characterization}, 
\textit{call for anti-group action} 
and \textit{undermining of fundamental narratives}.
For counter-hate, the thematic dimension
encodes three basic communicative aspects of hate which the counter speech addresses, labeled as:
\textit{the hate speaker}, \textit{the hate target} and \textit{the hate content}.

The rhetorical dimension, jointly defined for both hate and counter-hate, draws on Aristotle's theory on the means of persuasion, namely: Logos, Ethos and Pathos~\cite{aristotle1991rhetoric}. It includes four categories: \textit{emotive appeals} (inspired by Pathos), \textit{logical reasoning} (inspired by Logos), and the two credibility-based categories of \textit{positive self-construction} and \textit{negative representation of the other} (inspired by Ethos). 

We utilize this new scheme to annotate a sample of 92 complete X (formerly Twitter) conversations containing hateful speech towards Jews, with a total of 720 tweets. We perform several statistical analyses on this annotated sample, to demonstrate the potential in revealing the interactions between hate and counter-hate in the conversation taking place on social media. We further include \textit{like} counts for the annotated tweets to analyze the connections between the characteristics of hate expressions, the types of countering messages, and the amount of \textit{likes} received, demonstrating our scheme's potential in developing effective counter-hate strategies (in terms of maximizing the number of \textit{likes}).

To summarize, the paper makes the following contributions. Our novel annotation scheme offers a holistic theoretically driven approach for jointly annotating hate and counter-hate within organic social media conversations. The scheme decomposes hate/counter-hate into two separate dimensions, drawing a clear distinction between the content of the message and its form of expression. Using the scheme, we annotate a sample of complete social media conversations, and conduct several statistical analyses. While each dimension stands on its own right, their interactions unveil a wide array of insights into the dissemination of hate messages on social media and the diverse strategies employed to counter them.

The remainder of the paper is organized as follows: Section~\ref{sec:scheme} provides a comprehensive description of the annotation scheme. Section~\ref{sec:data} describes the annotation process, the resulting annotated sample, and presents some annotation examples. Section~\ref{subsec:data_analysis} details the statistical analyses conducted on the annotated sample and discusses the results. Section~\ref{sec:conclusion} provides a conclusion and discusses future work.

{\color{red} \textbf{Content Warning}: This document contains some examples of hateful content. This is strictly for the purpose of discussing the research presented here.
Please be aware that this content could be offensive and cause you distress.}

\section{Annotation Scheme}
\label{sec:scheme}

We present a novel annotation scheme that captures themes and rhetorical devices in social media conversations. The scheme jointly encodes hate and counter-hate messages along two distinct dimensions: \textit{thematic} and \textit{rhetorical}. 

The thematic categories are separately tailored for hate and counter-hate and were informed by extensive research on both~\cite{kaid2007encyclopedia,benesch2016counterspeech,davidson2017automated,chung-etal-2019-conan,arango-monnar-etal-2022-resources,khurana-etal-2022-hate,ron-etal-2023-factoring}, as well as by definitions utilized by international organizations.~\footnote{https://www.un.org/en/hate-speech/understanding-hate-speech/what-is-hate-speech.}
The rhetorical categories are based on the idea that, in the context of social media discourse, both hate and counter-hate messages are designed for persuasion. These categories were inspired by Aristotle's theory~\cite{aristotle1991rhetoric} and are shared by both hate and counter-hate messages.

Altogether, these two dimensions are designed to encompass a holistic view of hate and counter-hate speech on social media, as an interactive discursive situation in which participants engage in persuasion both to disseminate and to counteract hate messages.
While in this work the target group is \textit{Jews}, our annotation scheme is theoretically grounded and not based on any specific target-group attributes, rendering it applicable to any target population of interest. 

The annotation scheme is \textit{multi-labeled} in each of the two dimensions; i.e., a sample may be annotated with any subset of the thematic categories as well as any subset of the rhetorical categories.

\subsection{Thematic Dimension}
\label{subsec:thematic_categories}

\subsubsection{Hate}
\label{subsubsec:thematic_hate}

Hate messages are thematically encoded using three distinct categories. 
Theses categories encompass the ways in which the author perceives and addresses the target of hate: its characteristics (by expressing contempt towards the target), how it should be dealt with (by calling for anti-group action) and the illegitimacy of its fundamental narratives (by undermining these narratives). The categories are detailed below.

\begin{enumerate}
    \item \textbf{Contemptuous characterization} \textit{(contempt)}: speech that characterizes the target in a way that expresses intolerance, prejudice, and strong dislike towards the target, or which portrays the target as inferior (socially, morally, or in terms of competence). 
    
    \item \textbf{Call for anti-group action} (\textit{call for action}): speech that calls to take harmful, or potentially harmful action against the target, including the incitement of violence and discrimination.

    \item \textbf{Undermining of fundamental narratives} (\textit{under. narrative}): statements that belittle, mock, or deny a target’s fundamental narratives or core aspects or elements within them (as perceived by the group), which pertain to their history or sense of identity.
\end{enumerate}

\subsubsection{Counter-Hate}
\label{subsubsec:thematic_counter}

We categorize counter-hate messages into three thematic response types, emphasizing their communicative properties. 
These categories delineate the idea of reaction to hate statements by addressing three basic communicative aspects: the speaker (author of the hate message), the content of the hate message and the target of hate referred by the message. They reflect the counter-hate messenger's ability to counter different elements within their communication with hate messengers. The categories are described below. 

\begin{enumerate}
    \item \textbf{Addressing the hate-speaker} (\textit{speaker}): speech that opposes or denounces a hateful statement by explicitly appealing to its addresser (i.e., author), and/or referring to the addresser or the addresser's social group. 
    \item \textbf{Addressing the hate-target} (\textit{target}): speech that employs qualifications of the hated-target to oppose or denounce the hateful statement.
    \item \textbf{Addressing the hate-content} (\textit{content}): speech that opposes or denounces a hateful statement by referring to the substance of the hate-message. 
\end{enumerate}

While the annotation of hate rests upon the tweet itself, the annotation of counter hate speech can rely on contextual information outside of the annotated message, since it is, by definition, a response to a statement. 

\subsection{Rhetorical Dimension}
\label{subsec:rhetoric_categories}

Providing methods for examining rhetoric as a human phenomenon, Aristotle’s "On Rhetoric" is probably the most influential book on rhetoric in Western culture~\cite{aristotle1991rhetoric}. 
Inspired by 
Aristotle's theory, 
we designed four rhetorical categories to capture the way in which the hate/counter-hate message is communicated to the reader. Each of the categories was inspired by one of Aristotle's means of persuasion: Pathos -- an appeal to the emotions of the audience -- inspired the category of \textit{emotive appeals}, Logos -- an evocation of rational reasoning -- inspired the category of \textit{logical reasoning}, and Ethos -- substantiating the credibility of the speaker or deconstructing the credibility of the opponent -- inspired the two categories of \textit{positive self-construction}
and \textit{negative representation of the other}. 
All the rhetorical categories are shared by both hate and counter-hate messages. Below is a description of the categories.

\begin{enumerate}
    \item \textbf{Emotive appeals} (\textit{emotive}): speech that appeals to the emotions of its readers through its content (i.e., \textit{what} is said), which may also include an expression of emotions by the speaker; and/or its form of expression (i.e., \textit{how} is it said). 
    \item \textbf{Logical reasoning} (\textit{logical}): speech that is grounded in structural forms of argumentation and which evokes inferential (inductive or deductive) reasoning. Logical argumentation may also appeal to readers' rational thought while providing supportive evidence to their claims, as well as simply appeal to their common sense, which would not necessitate the presentation of external evidence. 
    \item \textbf{Positive self-construction} (\textit{pos. self}): speech that constructs the speaker's character as credible and trustworthy. 
    \item \textbf{Negative representation of the other} (\textit{neg. other}): speech that undermines the authority or credibility of the source/agent providing the information that the given message (hate or counter speech) speaks out against, or which generally presents the hated-target (or agents operating on its behalf) as not credible. This category was additionally inspired by subsequent work on Aristotle's Ethos~\cite{duthie2018classifying}.
\end{enumerate}

\section{Annotated Data Sample}
\label{sec:data}

In order to demonstrate the potential of our suggested encoding approach,
we set out to use it to annotate a sample of X (formerly Twitter) conversations. This section describes the annotation process and the annotated sample, presents some annotated examples and discusses them, and conducts several statistical analyses to reveal interesting insights derived from the data.

\begin{table}[!htbp]
\centering
\begin{tabular}{cccc}
\hline
\textbf{Hate} & \textbf{Counter} & \textbf{None} & \textbf{Total} \\
\hline
123 & 175 & 422 & 720 \\
\hline
\end{tabular}
\caption{Annotated data sample -- no. of tweets}
\label{tab:data_base_stats}
\end{table}

\begin{table}[!htbp]
\centering
\begin{tabular}{llc}
\hline
\multirow{3}{*}{\textbf{Thematic}} & Contempt & 101 \\
& Call for Action & 13 \\
& Under. Narrative & 44 \\
\hline
\multirow{4}{*}{\textbf{Rhetorical}} & Emotive & 101 \\
& Logical & 60 \\
& Pos. Self & 2 \\
& Neg. Other & 3 \\
\hline
\end{tabular}
\caption{Annotation statistics: hate}
\label{tab:base_stats_hate}
\end{table}

\begin{table}[!htbp]
\centering
\begin{tabular}{llc}
\hline
\multirow{3}{*}{\textbf{Thematic}} & Speaker & 125 \\
& Target & 30 \\
& Content & 137 \\
\hline
\multirow{4}{*}{\textbf{Rhetorical}} & Emotive & 106 \\
& Logical & 101 \\
& Pos. Self & 3 \\
& Neg. Other & 69 \\
\hline
\end{tabular}
\caption{Annotation statistics: counter-hate}
\label{tab:base_stats_counter}
\end{table}

\begin{table*}[!htbp]
\centering
\begin{tabular}{lcccc}
\hline
& \textbf{Emotive} & \textbf{Logical} & \textbf{Pos. Self} & \textbf{Neg. Other} \\
\hline
\textbf{Contempt} & 92 & 45 & 0 & 3  \\
\textbf{Call for Action} & 12 & 9 & 0 & 1 \\
\textbf{Under. Narrative} & 26 & 27 & 2 & 1 \\
\hline
\end{tabular}
\caption{Hate tweet counts: thematic vs. rhetorical}
\label{tab:hate_cross_stats}
\end{table*}

\begin{table*}[!htbp]
\centering
\begin{tabular}{lcccc}
\hline
& \textbf{Emotive} & \textbf{Logical} & \textbf{Pos. Self} & \textbf{Neg. Other} \\
\hline
\textbf{Speaker} & 79 & 75 & 1 & 65  \\
\textbf{Target} & 16 & 20 & 3 & 8 \\
\textbf{Content} & 78 & 90 & 3 & 41 \\
\hline
\end{tabular}
\caption{Counter-hate tweet counts: thematic vs. rhetorical}
\label{tab:counter_cross_stats}
\end{table*}

\subsection{Description}
\label{subsec:data_description}

We used the data collected by~\citet{ron-etal-2023-factoring} -- a corpus of complete raw X conversations (with corresponding meta-data, such as retweets, replies, likes), 
extracted through Twitter API v2 using a set of 
filter words pertaining to Jews and Judaism~\footnote{Jews as the target group include Zionist-related references.} (to maximize the chance of encountering Jew-related hate speech). We extracted a sample of conversations exhibiting both hate and counter-hate speech, consisting of 92 conversations and a total of 720 tweets, posted between August 26th, 2018 and January 22nd, 2023.

The sample was annotated by an expert annotator (who underwent dedicated training for this task) according to the definitions described in Section~\ref{sec:scheme}. Each conversation was annotated in a ``depth-first search'' manner, i.e., beginning with the root tweet, and following each sub-branch of the conversation to its end before moving to the next sub-branch. Each tweet in the conversation was encoded either with a subset of the three hate thematic categories (Section~\ref{subsubsec:thematic_hate}), a subset of the three thematic counter-hate categories (Section~\ref{subsubsec:thematic_counter}) or `none' if no hate or counter-hate expressions were detected. If the tweet was encoded with any thematic category, it was also encoded with a subset of the four rhetorical categories (Section~\ref{subsec:rhetoric_categories}). Tables~\ref{tab:data_base_stats}, ~\ref{tab:base_stats_hate} and ~\ref{tab:base_stats_counter} summarize basic statistics for the annotated sample.

\subsection{Examples}
\label{subsec:examples}

The first example presents a hate tweet that refers to Jews in a demeaning manner and expresses strong disliking of them. It also constitutes a degrading characterization of the hated group that inserts a sense of ridicule on the one hand, and aversion on the other. The counter-tweet responds with name-calling the hate tweet's author, referring to them as racist, hence also labeling the statement as such and addressing the hate-content. 

\ex<example1> \textbf{Hate:} "@username I am a proud anti Synagogue of Satan traditional Catholic who says that proudly from Brooklyn, NY. If you don't like that I could care a less. \#CatholicTwitter @username @username"

\textbf{Thematic:} \textit{contempt} \vspace{0pt}

\textbf{Rhetorical:} \textit{emotive} \\

\textbf{Counter:} "@username @username @username @username What you are is a weirdo who has been reported for racism."

\textbf{Thematic:} \textit{speaker, content} \vspace{0pt}

\textbf{Rhetorical:} \textit{emotive}

\xe

In the next example, the hate tweet refers to the Holocaust as "Holohoax", a term known to be employed by Holocaust deniers, therefore constituting an undermining of a fundamental Jewish historical narrative. The tweet claims that it is supported by evidence, an attempt to advance the statement through logical argumentation.

The counter-response directly addresses the accusation, i.e., the content, by referencing evidence that disproves the claim. It also attacks the speaker, discrediting their credibility, and substantiates its counter claim by relating to a relevant description of the hated target. The counter tweet additionally alludes to an unjust treatment of the target, referencing another population that had suffered.

\ex<example2> \textbf{Hate:} "Holohoax evidence keeps pouring in... <<URL>>"

\textbf{Thematic:} \textit{under. narrative} \vspace{0pt}

\textbf{Rhetorical:} \textit{logical} \\

\textbf{Counter:} "@username this debunked years ago. Yr a liar. There are many places that hold authentic documents not Irvings crap. No one argues about 20 million Soviet dead. Or any other nationality. The Nazis admitted it, people witnessed it and Jewish pop. still not recovered from pre ww2 numbers"

\textbf{Thematic:} \textit{speaker, target, content} \vspace{0pt}

\textbf{Rhetorical:} \textit{emotive, logical, neg. other}

\xe

The third example depicts a hateful tweet that employs racial slurs and derogatory language to reference the hated target, while claiming that some truth was exposed. 
The first counter tweet plainly states that the hateful tweet employs a racist term, then directly addresses the speaker with animosity. The second counter tweet addresses the speaker with hostility, while additionally undermining their credibility and the truthfulness of their statement.

\ex<example3> \textbf{Hate:} "@username Exposed facts are a bitch to a Ziotroll. <<URL>> US-led Wars for this Israeli plan <<URL>> cost Americans thousands of warrior's lives while US cities and infrastructure rot. Zios are America's internal enemy, including Christian Zios"

\textbf{Thematic:} \textit{contempt} \vspace{0pt}

\textbf{Rhetorical:} \textit{emotive, logical} \\

\textbf{Counter \#1:} "@username Zio is a racist term U r an antisemitic neo nazi Now crawl back under ur rock"

\textbf{Thematic:} \textit{speaker, content} \vspace{0pt}

\textbf{Rhetorical:} \textit{emotive, logical} \\

\textbf{Counter \#2:} "@username Your diatribe evidences you know nothing about Israel, Jews and Judaism. You merely parrot propaganda with fabricated b*llshit, delusional myths, concocted stories \& fake facts."

\textbf{Thematic:} \textit{speaker, content} \vspace{0pt}

\textbf{Rhetorical:} \textit{emotive, logical, neg. other}

\xe

\begin{table*}[!htbp]
\centering
\begin{tabular}{r|l|ccc}
& & \multicolumn{3}{c}{\textbf{Counter}} \\
\hline
& & \textbf{Speaker} & \textbf{Target} & \textbf{Content} \\
\hline
\multirow{3}{*}{\rotatebox[origin=c]{90}{\textbf{Hate}}} & \textbf{Contempt} & 1.02 & 0.25 & 1.12 \\
& \textbf{Call for Action} & 1.31 & 0.31 & 1.08 \\
& \textbf{Under. Narrative} & 1.05 & 0.20 & 1.20 \\
\hline
\end{tabular}
\caption{Average number of counter-hate tweets in response to a hate tweet, over the thematic categories}
\label{tab:thematic_cross_statistics}
\end{table*}

\begin{table*}[!htbp]
\centering
\begin{tabular}{r|l|cccc}
& & \multicolumn{4}{c}{\textbf{Counter}} \\
\hline
& & \textbf{Emotive} & \textbf{Logical} & \textbf{Pos. Self} & \textbf{Neg. Other} \\
\hline
\multirow{4}{*}{\rotatebox[origin=c]{90}{\textbf{Hate}}} & \textbf{Emotive} & 0.77 & 0.76 & 0.01 & 0.51 \\
& \textbf{Logical} & 0.73 & 0.98 & 0.03 & 0.57 \\
& \textbf{Pos. Self} & 1.50 & 1.00 & 0.00 & 1.00 \\
& \textbf{Neg. Other} &  1.00 & 1.33 & 0.00 & 0.33 \\
\hline
\end{tabular}
\caption{Average number of counter-hate tweets in response to a hate tweet, over the rhetorical categories}
\label{tab:rhetoric_cross_statistics}
\end{table*}

\section{Statistical Analysis}
\label{subsec:data_analysis}

In this section, we detail several statistical analyses performed over our annotated data sample (along with additional meta-data), to demonstrate the potential of our encoding approach to provide insights into the relationship between hate and counter-hate expressions on social media, and its possible use in devising effective strategies to counter online hate.

\subsection{Tweet Counts}
\label{subsubsec:tweet_counts}

Tables~\ref{tab:hate_cross_stats} and~\ref{tab:counter_cross_stats} summarize cross-statistic for thematic vs. rhetorical categories in hate and counter hate tweets; for example, the number of tweets that undermine the narrative of the target and contain logical reasoning is 27. Table~\ref{tab:hate_cross_stats} reveals that hate speech in our annotated sample focuses mainly on contemptuous expressions towards the target, with a smaller number of tweets undermining its narrative and even fewer tweets calling for anti-group action. In terms of rhetoric, hate tweets mainly utilize emotive \& logical characteristics, where contemptuous tweets tend to be more on the emotive side and the other thematic categories are equally divided between emotion and logic. On the other hand, Table~\ref{tab:counter_cross_stats} shows that in addition to their extensive use of emotive appeals and logical reasoning, counter-hate tweets also rely on negative representation of the other as a rhetorical instrument. In rhetorical terms, they mainly address the speaker and the content (while occasionally addressing the target) of the hate tweet.

Table~\ref{tab:thematic_cross_statistics} displays the average number of response counter-hate tweets to a hate tweet, over the thematic categories; for example, the average number of counter-hate tweets addressing the speaker in response to a tweet containing a call for anti-group action is 1.31. Evidently, hate tweets received more speaker-addressed and content-addressed counter-hate responses than target-addressed ones. Contemptuous and narrative-undermining hate tweets received slightly more content-addressed counters, while call for anti-group action received slightly more speaker-addressed counters. Overall, hateful tweets calling for anti-group action received more counter-hate responses than the other hate categories.

Table~\ref{tab:rhetoric_cross_statistics} shows the average number of counter-hate response tweets to a hate tweet, over the rhetorical categories; for example, the average number of counter-hate logical responses to an emotive hate tweet is 0.76. While emotive hate speech received (on average) a similar amount of emotive and logical counter responses, logical hate speech received more logical counter responses than emotive ones. Positive self-constructive hate tweets attracted more emotive responses, while hate expressions depicting negative representation of the other received more logical counter responses. Interestingly, positive representation of the other is virtually never used in counter-hate responses, possibly pointing to it being conceived as an ineffective rhetorical means to counter hate speech.

\subsection{\textit{Like} Counts}
\label{subsubsec:like_counts}

The number of \textit{likes} a tweet receives may be considered a proxy to the degree of impact of the tweet within the social media platform~\cite{mathew2019thou}. The corpus collected by~\citet{ron-etal-2023-factoring} contains public metrics for each tweet, allowing us to perform an analysis which takes into account the number of \textit{likes} received by each hate and counter-hate tweet in our annotated sample. 

\begin{table*}[!htbp]
\centering
\begin{tabular}{lcccc}
\hline
& \textbf{Emotive} & \textbf{Logical} & \textbf{Pos. Self} & \textbf{Neg. Other} \\
\hline
\textbf{Contempt} & 22.17 & 30.67 & 0.00 & 0.00  \\
\textbf{Call for Action} & 3.08 & 5.44 & 0.00 & 0.00 \\
\textbf{Under. Narrative} & 5.73 & 31.59 & 0.50 & 0.00 \\
\hline
\end{tabular}
\caption{Average number of \textit{like}s received by hate tweets: thematic vs. rhetorical}
\label{tab:hate_cross_likes}
\end{table*}

\begin{table*}[!htbp]
\centering
\begin{tabular}{lcccc}
\hline
& \textbf{Emotive} & \textbf{Logical} & \textbf{Pos. Self} & \textbf{Neg. Other} \\
\hline
\textbf{Speaker} & 0.75 & 0.43 & 1.00 & 0.71 \\
\textbf{Target} & 0.69 & 0.20 & 0.33 & 0.88 \\
\textbf{Content} & 1.31 & 0.49 & 0.33 & 0.71 \\
\hline
\end{tabular}
\caption{Average number of \textit{like}s received by counter-hate tweets: thematic vs. rhetorical}
\label{tab:counter_cross_likes}
\end{table*}

Table~\ref{tab:hate_cross_likes} details the average number of \textit{likes} received by a hateful tweet in our annotated sample, divided by thematic and rhetorical categories, while Table~\ref{tab:counter_cross_likes} displays the same for counter-hate tweets. 
First, we can observe that hate tweets attracted a significantly higher number of \textit{likes} than counter-hate tweets, stressing the notion that negative discourse is the bread and butter of social media. Among the three thematic hate categories, contemptuous tweets received the highest average number of \textit{likes} (by a large margin), followed by narrative undermining tweets and tweets calling for anti-group action, signaling that users find it easier to support general contempt towards the target rather than an actual call for action against it. Interestingly, logical hateful expressions received notably more \textit{likes} than emotive ones, while the opposite is true for counter-hate tweets, where emotive tweets received more \textit{likes} than logical ones, indicating that users prefer logical expressions of hate, but emotional attempts to counter it. 

Table~\ref{tab:thematic_cross_likes} shows the average number of \textit{likes} received by counter-hate tweets, conditioned on the thematic categories of the counter-hate tweet as well as the hate tweet to which it responded. For example, tweets addressing the speaker, posted in order to counter a hate tweet containing a call for anti-group action, received an average of 0.71 \textit{likes}. 
Overall, addressing the target of the hateful tweet yielded the lowest average number of \textit{likes} among the counter-hate thematic categories. 
Addressing the content of the hateful message seems to produce the highest average number of \textit{likes} when countering contemptuous or call-for-action tweets, while addressing the speaker attracted the highest average number of \textit{likes} when countering hateful expressions undermining the target group's narratives (although by a small margin). 

Table~\ref{tab:rhetoric_cross_likes} displays a similar analysis performed on the rhetorical dimension of the annotated tweets. We observe that emotive counter-hate garnered the highest average number of \textit{likes} (by a large margin) when countering an emotive hateful tweet, compared to the other rhetorical categories. Countering logical hate expressions, however, received a relatively more similar number of \textit{likes} when using each of the four rhetorical categories, with emotive and negative representation of the other counters in the lead and logical and positive self-construction not far behind. Emotive and logical counter messages achieved the largest number of \textit{likes} when used to counter positive self-constructive hate tweets, while countering negative representation of the other yielded no \textit{likes} whatsoever, regardless of the rhetorical category used.

This type of analysis allows us to not only explore the distribution of received \textit{likes} over the different combinations of thematic-categories, but also holds the potential to help in devising a strategy to counter hateful messages by choosing the most (empirically) effective type of response according to our scheme, in terms of maximizing the number of \textit{likes}. However, this would entail annotating a larger, more diverse dataset, and possibly analyzing higher-order relations between the thematic and rhetorical categories (e.g., combinations of several thematic categories with several rhetorical ones, leveraging the multi-labeled nature of our annotation scheme). 

\begin{table*}[!htbp]
\centering
\begin{tabular}{r|l|ccc}
& & \multicolumn{3}{c}{\textbf{Counter}} \\
\hline
& & \textbf{Speaker} & \textbf{Target} & \textbf{Content} \\
\hline
\multirow{3}{*}{\rotatebox[origin=c]{90}{\textbf{Hate}}} & \textbf{Contempt} & 0.66 & 0.52 & 1.01 \\
& \textbf{Call for Action} & 0.71 & 0.75 & 0.79 \\
& \textbf{Under. Narrative} & 0.65 & 0.22 & 0.60 \\
\hline
\end{tabular}
\caption{Average number of \textit{like}s received by a counter-hate tweet in response to a hate tweet -- thematic categories analysis}
\label{tab:thematic_cross_likes}
\end{table*}

\begin{table*}[!htbp]
\centering
\begin{tabular}{r|l|cccc}
& & \multicolumn{4}{c}{\textbf{Counter}} \\
\hline
& & \textbf{Emotive} & \textbf{Logical} & \textbf{Pos. Self} & \textbf{Neg. Other} \\
\hline
\multirow{4}{*}{\rotatebox[origin=c]{90}{\textbf{Hate}}} & \textbf{Emotive} & 1.13 & 0.22 & 0.00 & 0.44 \\
& \textbf{Logical} & 0.80 & 0.63 & 0.50 & 0.79 \\
& \textbf{Pos. Self} & 2.00	& 2.00 & 0.00 & 1.50 \\
& \textbf{Neg. Other} & 0.00 & 0.00 & 0.00 & 0.00 \\
\hline
\end{tabular}
\caption{Average number of \textit{like}s received by counter-hate tweets in response to a hate tweet-- rhetorical categories analysis}
\label{tab:rhetoric_cross_likes}
\end{table*}

\section{Conclusion}
\label{sec:conclusion}

In light of the proliferation of hate speech on social media, we introduce a novel two-dimensional (thematic and rhetorical) multi-labeled scheme for hate and counter-hate annotation. 
The thematic dimension categorizes hate speech based on the messenger's perception of the target of hate, while counter-hate speech is categorized based on the communicative properties of the response.
The rhetorical dimension encodes \textbf{how} hate and counter-hate messages are communicated, drawing upon Aristotle's classical means of persuasion - Logos, Ethos, and Pathos.

Our annotation scheme presents a holistic framework for jointly annotating hate speech and counter-hate speech within organic social media conversations. Breaking down hate and counter-hate into two distinct dimensions, the scheme highlights the disparity between message content and expression of persuasive related form. Consequently, 
it holds the potential to reveal the rhetorical mechanisms of hate and counter-hate dissemination on social media.

To showcase the efficacy of our scheme, we 
annotate real-world samples of conversations from X (formerly Twitter), illustrating instances of both hate speech and counter-hate speech directed toward the targeted group. The findings reveal promising insights into potential patterns of hate and counter-hate dissemination. For instance, we observed that tweets advocating for anti-group actions generally elicited more counter-hate responses compared to other categories of hate speech (contemptuous characterization and undermining of fundamental narratives). Additionally, hate tweets tended to elicit more logical counter-hate rhetoric on average than an emotional one.

Another analysis takes into account the number of \textit{likes} received by tweet, which may be considered as proxy to the tweet's impact within the platform.
Notably, hate tweets garnered a significantly higher number of \textit{likes} than counter-hate tweets. We also noted an intriguing trend where hate speech with logical rhetoric received notably more \textit{likes} than emotive ones. Conversely, for counter-hate tweets, emotive expressions tended to garner more \textit{likes} than logical ones. We speculate that this pattern may suggest a preference among users for logical expressions of hate, but emotional attempts to counter it.

We believe that this preliminary work demonstrates our scheme's potential in paving the way for a diverse array of both theoretical and practical inquiries into the spread of hate messages and the effectiveness of their countering. 

In continuation of this work, we are currently engaged in an effort to construct a large annotated dataset, using a team of trained annotators, and validated via inter-annotator reliability tests. This dataset will be used to train a supervised prediction model, which in turn will allow us to automatically annotate vast amounts of data and perform analyses in a substantially larger scale. In addition, we also intend to expand our project to Muslim-directed hate and counter-hate, in order to validate the ability of our annotation scheme to generalize to another target group. 

\section*{Limitations}
\label{sec:limitations}

One apparent limitation of our work is the limited size of the annotated data sample.
As mentioned in Section~\ref{sec:conclusion}, we have already recruited and trained a group of annotators and have begun the process of expanding the sample into a larger annotated dataset.

Another limitation of this work lies in the fact that the annotation scheme was applied to a single target population. 
Since our annotation scheme is generic, 
this can be remedied by using it to annotate hate and counter-hate directed towards additional target groups;
as stated in Section~\ref{sec:conclusion}, we intend to expand our project to Muslim-directed hate and counter-hate.

Finally, the scheme is limited in the sense that it only annotates direct expressions of hate and counter-hate speech, as opposed to messages that perpetrate them through expression of agreement or support. This may be solved by expanding the scheme to encompass supportive and non-supportive engagement with both hate and counter speech.

\section*{Ethics Statement}

Online hate speech has dangerous real-world consequences~\cite{siegel2020online}. This work utilizes a novel annotation scheme to analyze social media posts, potentially containing hateful expressions, in terms of various discursive forms of communication. While we intend the annotation scheme to be used exclusively for academic research as well as for potentially improving counter hate speech interventions in online conversations, it could potentially be misused to produce more compelling online hateful speech. We defiantly oppose such potential use and strongly believe in our work's value in propagating tolerant and peaceful discourse on social media.

\bibliography{arxiv,anthology}

\end{document}